%% file: main.tex
\documentclass{article}
\usepackage{spconf,amsmath,amssymb, graphicx,mathtools}

\title{Using Deep Image Priors to Generate Counterfactual Explanations}
\name{Vivek Narayanaswamy$^1$\thanks{This work was supported in part by the ASU SenSIP Center, Arizona State University. Portions of this work was performed under the auspices of the U.S. Department of Energy by Lawrence Livermore National Laboratory under Contract DE-AC52-07NA27344.}, Jayaraman J. Thiagarajan$^2$, Andreas Spanias$^1$}
\address{$^1$ Arizona State University, $^2$ Lawrence Livermore National Laboratory}
\begin{document}
%
\maketitle
\begin{abstract}

\input{abstract.tex}

\end{abstract}
\begin{keywords}
Deep Image Prior, pre-images, counterfactual reasoning, black-box models, explainable AI
\end{keywords}
\section{Introduction}
\label{sec:intro}

\input{intro.tex}

\section{Pre-Image Recovery in Neural Networks}
\label{sec:background}
\input{background}

\section{Proposed Approach}
\label{sec:approach}

\input{approach.tex}

\section{Experiments}
\label{sec:results}

\input{results.tex}

\section{Conclusion}
\label{sec:conclusion}

\input{conclusion.tex}
\nocite{thiagarajan2020calibrating}

\bibliographystyle{IEEEbib}
\bibliography{ref}

\end{document}

%% file: abstract.tex
Through the use of carefully tailored convolutional neural network architectures, a deep image prior (DIP) can be used to obtain pre-images from latent representation encodings. Though DIP inversion has been known to be superior to conventional regularized inversion strategies such as total variation, such an over-parameterized generator is able to effectively reconstruct even images that are not in the original data distribution. This limitation makes it challenging to utilize such priors for tasks such as counterfactual reasoning, wherein the goal is to generate small, interpretable changes to an image that systematically leads to changes in the model prediction. To this end, we propose a novel regularization strategy based on an auxiliary loss estimator jointly trained with the predictor, which efficiently guides the prior to recover natural pre-images. Our empirical studies with a real-world ISIC skin lesion detection problem clearly evidence the effectiveness of the proposed approach in synthesizing meaningful counterfactuals. In comparison, we find that the standard DIP inversion often proposes visually imperceptible perturbations to irrelevant parts of the image, thus providing no additional insights into the model behavior.


%% file: intro.tex
The \textit{Deep Image Prior} (DIP)~\cite{ulyanov2018deep}, which leverages the structure of an untrained, carefully tailored convolutional neural network to generate images, has been widely adopted to solve a variety of ill-posed restoration tasks in computer vision. In contrast to approaches that use priors based on pre-trained generative models (e.g, Generative Adversarial Networks (GANs))~\cite{anirudh2020mimicgan, shah2018solving, yeh2017semantic, asim2018blind,narayanaswamy2020unsupervised, goodfellow2014generative} to solve inverse problems, DIP produces semantically meaningful reconstructions without any knowledge of the image manifold. 

A natural application for DIP is to effectively regularize the problem of inverting the encoded image representation, from an arbitrary layer of a \textit{black-box} neural network, to recover the original image. Commonly referred to as ``pre-image'' recovery~\cite{ulyanov2018deep, mahendran2015understanding}, it is a critical diagnostic tool to study the invariances captured by a neural network. Classical approaches that rely on statistical priors, e.g., total variation~\cite{mahendran2015understanding} or learned priors for every layer~\cite{dosovitskiy2016inverting} are known to be insufficient to recover images from their encodings. In contrast, DIP has been found to produce high-quality reconstructions, particularly when inverted from early layers of the network. Despite its effectiveness, due to the inherent lack of knowledge about the underlying data manifold, the over-parameterized generative network in DIP can recover even out-of-distribution (OOD) pre-images. This is a crucial bottleneck when utilizing DIP to produce counterfactual explanations for interpreting predictive models -- the changes required to the given image that will lead to a specific change in its corresponding prediction. As we will show in this paper, counterfactual explanations obtained using DIP-based pre-image recovery most often corresponds to adversarial examples, wherein visually imperceptible changes are proposed to irrelevant parts of the image, and hence cannot be used to gain insights into the model behavior.

When a pre-trained generative model is available, one can always reproject an OOD sample onto the image manifold, e.g., MimicGAN~\cite{anirudh2020mimicgan}, to produce meaningful reconstructions. However, in this paper, we consider the setup where we assume no access to the original data (or a generative model), but only to a \textit{black-box} predictive model. We propose a novel regularization to the DIP pre-image recovery process, based on an auxiliary \textit{loss estimation} network that is trained alongside the predictive model. Recent studies on active learning and explainable AI~\cite{yoo2019learning, thiagarajan2020accurate} have found that such a loss estimator can capture the inherent model uncertainties, and more importantly detect challenging distribution shifts. 

\noindent
\textbf{Proposed Work}:~In this paper, we propose to utilize gradients from the loss estimator to obtain semantically meaningful pre-images using DIP, even when the latent space encoding corresponds to an OOD image. Interestingly, we show that such a regularization is essential for producing realistic counterfactuals from the underlying data manifold. For any user-specified hypothesis, the loss estimator provides useful gradients to introduce interpretable changes to the pre-image as the prediction changes to the specified class label. Through extensive analysis on a real world dataset of lesion images from the ISIC 2018 challenge~\cite{codella2019skin, tschandl2018ham10000}, we show that the proposed approach produces significantly superior pre-images when compared to the standard DIP inversion.

%% file: background.tex
Pre-image recovery~\cite{mahendran2015understanding, mahendran2016visualizing} refers to the ill-posed problem of inverting an arbitrarily encoded representation to a realization on the (unknown) image manifold. Let $\mathbf{x}_{0}$ denote an image and $\Psi(.)$ be any differentiable layer of a neural network. The pre-image recovery can be mathematically formulated as
\begin{equation}
\mathbf{x}^{*} =arg\min_{\mathbf{x} \in \mathbb{R}^{H\times W\times C}} \mathcal{L}(\Psi(\mathbf{x}), \Psi(\mathbf{x}_{0}))  + \lambda\mathcal{R}(\mathbf{x}).
\label{generalinv}
\end{equation}Here $\mathcal{L}$ is usually the L2 norm between the encoded representations for the true image and the estimated pre-image, and $\mathcal{R}(.)$ is a suitable regularizer. Conventional approaches use explicit regularizers such as the $\alpha-$norm of the images~\cite{mahendran2015understanding} or total variation, which enforces piece wise consistency, to tractably solve this inversion problem. Recently, Ulyanov \textit{et al.} proposed the Deep Image Prior~\cite{ulyanov2018deep} and showed that using the neural network structure as an implicit regularizer (e.g., a U-Net architecture~\cite{ronneberger2015u} in our study) can lead to significantly superior pre-image recovery. Following this work, we develop a DIP-based inversion method to generate interpretable, counterfactual explanations for a predictive model.

%% file: approach.tex
\begin{figure}
    \centering
    \includegraphics[width = \columnwidth]{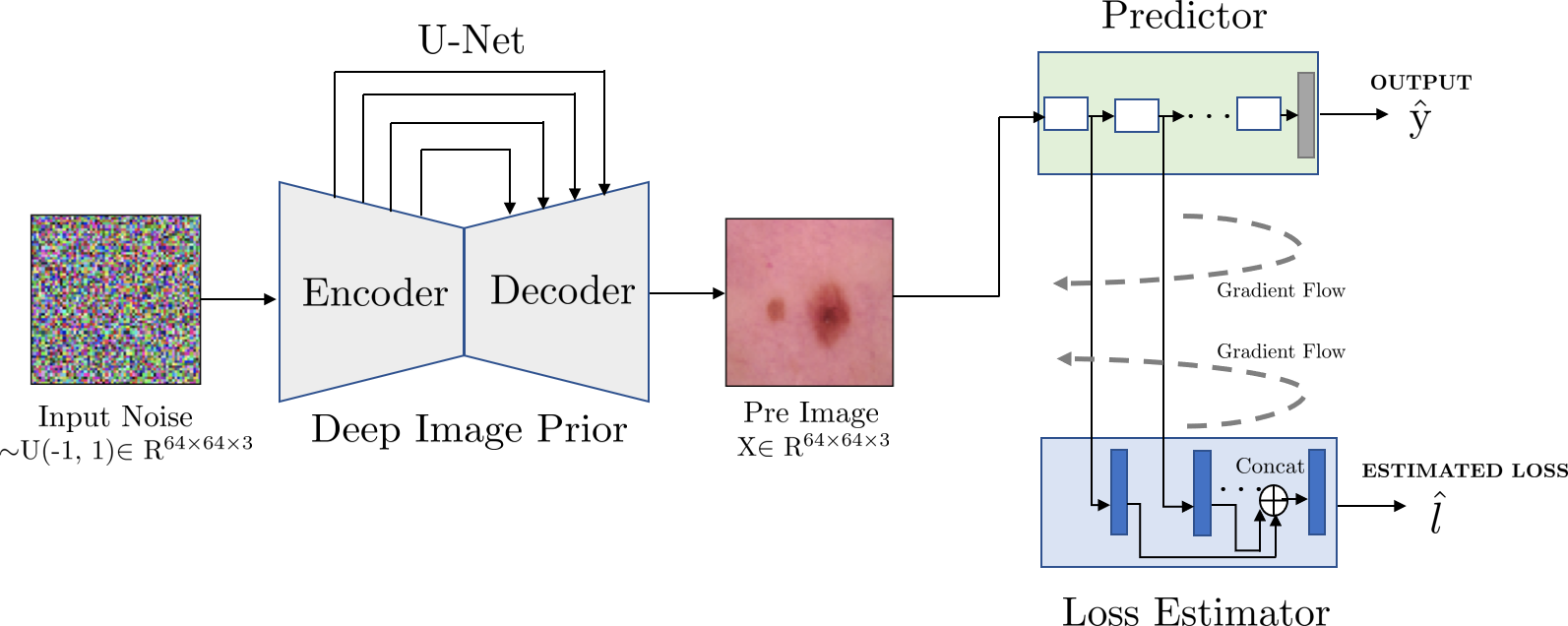}
    \caption{An illustration of the proposed pre-image recovery approach based on Deep Image Prior.}
    \label{fig:blk_diagram}
\end{figure}

In this section, we describe the proposed approach for using DIP to obtain pre-images from underlying data distribution, which in turn can be used for generating counterfactual explanations. The key idea of our approach (see Figure \ref{fig:blk_diagram}) is to utilize an auxiliary loss estimator, trained alongside the predictor, to regularize the DIP inversion process. 

\subsection{Loss Estimator Design}
Let us denote a multi-class classifier that takes as input an image $\mathbf{x} \in \mathbb{R}^{H\times W\times C}$ to predict the output label $\mathrm{y} \in \mathcal{Y} \coloneqq [1,\cdots,K]$ as $\mathcal{F}_{\Theta}(\mathbf{x})$. Here, $\mathbf{x}$ is a $C$ channel image of height $H$ and width $W$, $\Theta$ denotes the parameters of the model and $K$ represents the total number of classes.  
For the given training data $\{(\mathbf{x}_i,\mathrm{y}_i)\}_{i=1}^N$, where $N$ denotes the number of samples, we learn the parameters $\Theta$ using the standard cross-entropy loss function $\mathcal{L}_{pri}(\mathrm{y}, \hat{\mathrm{y}})$. Here, $\mathcal{L}_{pri}$ measures the error between the true and predicted class labels. In order to capture the inherent model uncertainties and semantics of the data manifold, we construct an auxiliary model $\mathcal{G}_{\Phi, \Theta }(\mathbf{x})$ that learns to estimate the loss $\ell = \mathcal{L}_{pri}(\mathrm{y}, \hat{\mathrm{y}})$. As depicted in Figure \ref{fig:blk_diagram}, the loss estimator uses the hidden representations extracted from every convolutional block (can correspond to one or more convolutional layers, non-linear activation and normalization) of the predictor (4 blocks in our case). We utilize a linear layer with a ReLU activation to transform each hidden representation from the predictor and finally concatenate them to produce the loss estimate $\hat{\ell}$. Similar to~\cite{yoo2019learning, thiagarajan2020accurate}, we utilize an auxiliary loss function $\mathcal{L}_{aux}(\ell, \hat{\ell})$ to train the parameters $\Phi$ of the loss estimator. In particular, we adopt the contrastive loss which aims to preserve the ordering of the samples based on their corresponding losses from the predictor. More specifically, we ensure that the relative change of sign in the loss values between the samples of a batch from the primary loss function is captured by the auxiliary network. Let $\ell_i$ and $\ell_j$ denote the losses of samples $\mathbf{x}_i$ and $\mathbf{x}_j$, while the corresponding estimates from $\mathcal{G}$ are $\hat{\ell}_i$ and $\hat{\ell}_j$ respectively. Mathematically, 
\begin{align}
    \mathcal{L}_{aux} = &\sum_{(i,j)}\max \bigg(0, -\mathbb{I}(\ell_i,\ell_j) . (\hat{\ell}_i - \hat{\ell}_j) + \gamma \bigg), \\
    &\nonumber \text{where } \mathbb{I}(\ell_i,\ell_j) = \begin{cases}
1 &\text{if $\ell_i > \ell_j$},\\
0 &\text{otherwise}.
\end{cases}
    \label{eqn:laux1}
\end{align}Here $\gamma$ is an optional margin parameter ($\gamma$ = 1 for our experiments). Interestingly, in~\cite{thiagarajan2020accurate}, it was showed that the loss estimator can effectively detect distribution shifts. The overall objective for the joint optimization of the predictor and loss estimator is given by
\begin{equation}
    \mathcal{L}_{total} =\beta_{1} \mathcal{L}_{pri} + \beta_{2} \mathcal{L}_{aux}. 
\end{equation}In our experiments, we used $\beta_{1}$=1 and $\beta_{2}$=0.5. 

\begin{figure*}[t]
    \centering
    \includegraphics[width = 0.99\textwidth]{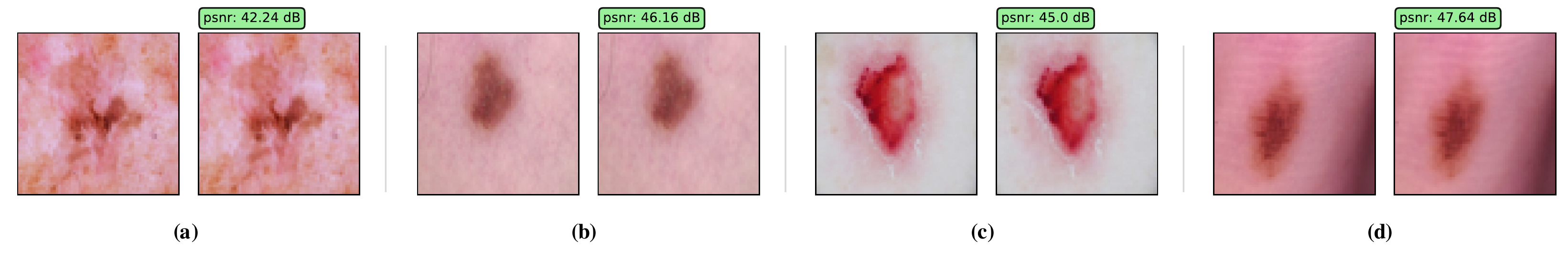}
    \caption{\textbf{Pre-image recovery using DIP}. (a)-(d) As reported in~\cite{ulyanov2018deep}, the Deep Image Prior can effectively invert deep representations of samples on the true image manifold. Including the proposed regularization provides only marginal improvements.}
    \label{fig:exp1}
\end{figure*}

\begin{figure*}[t]
    \centering
    \includegraphics[width = 0.99\textwidth]{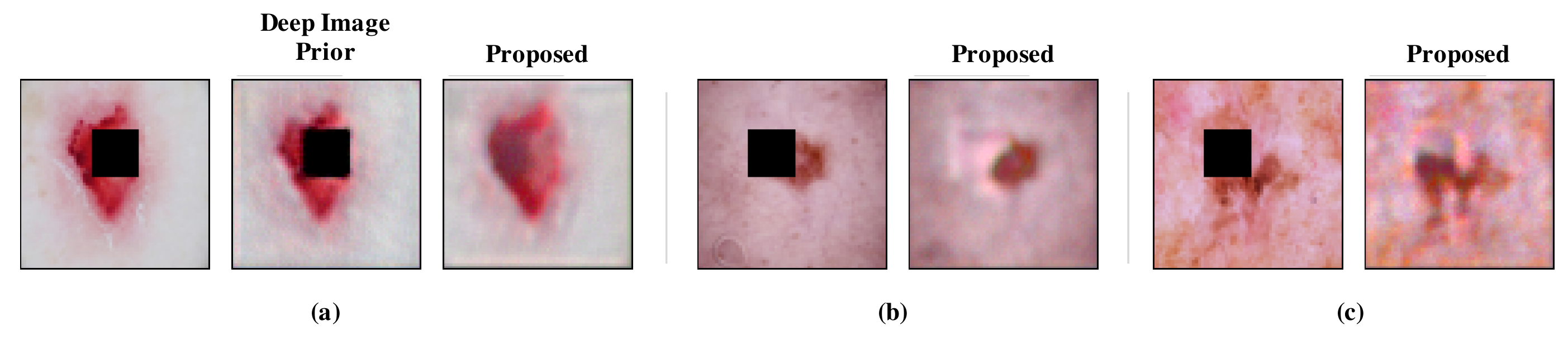}
    \caption{\textbf{Pre-image recovery for out-of-distribution images}. (a)-(c) The standard DIP overfits even to the cropped image, while the proposed approach guides DIP to produce more realistic images without an explicit data prior. 
    }
    \label{fig:exp2}
\end{figure*}

\subsection{Regularizing Pre-image Recovery using DIP}
In~\cite{ulyanov2018deep}, it was showed that a DIP regularizes the pre-image inversion problem to accurately recover images without any pre-training. However, due to the inherent lack of data priors, the formulation in \eqref{generalinv} can reconstruct even images that do not belong to the original data distribution, which makes counterfactual generation very challenging. In order to handle this issue, we propose to utilize the pre-trained loss estimator to further guide the inversion. Loosely speaking, the losses from the loss estimator are reflective of whether a given sample is from the underlying data distribution. For example, a significantly higher loss for a sample can imply that the model is less confident about the prediction or the sample is out-of-distribution. By enforcing the loss estimator to operate under high confidence regimes (i.e., low loss estimates), one can obtain meaningful gradients to generate images that belong to the original distribution. Formally, 
\begin{align}
{\mathbf{x}}^{*} =arg\min_{\mathbf{x} \in \mathbb{R}^{H\times W\times C}} \mathcal{L}(\Psi(\mathbf{x}), \Psi(\mathbf{x}_{0}))  +\lambda_{1}\mathcal{M}(\mathbf{x}).
\label{generalinv_le}
\end{align}where $\mathcal{M}(\mathbf{x})$ is the new regularization based on the loss estimator. In our experiments, we utilize the $L2$ norm between the estimated loss and a target loss indicating the desired confidence level in the prediction. 

A natural extension to this formulation allows the generation of meaningful counterfactuals defined on the original image manifold. Counterfactual reasoning is a critical tool in explaining the predictions of a classifier. Given a hypothesis about a sample, counterfactual evidences seek to systematically add perturbations to move the sample towards that hypothesis. More specifically, we extend the pre-image generation process in Eq. \eqref{generalinv_le} to produce a counterfactual by including a weighted conditional entropy loss from the black-box model. While the conditional entropy term produces the required gradients to alter the current hypothesis of a sample, the loss estimator produces the gradients to produce an image on the original manifold:
\begin{align}
\nonumber{\mathbf{x}}^{*} =arg\min_{\mathbf{x} \in \mathbb{R}^{H\times W\times C}} \mathcal{L}(\Psi(\mathbf{x}), \Psi(\mathbf{x}_{0}))    +\lambda_{1}\mathcal{M}(\mathbf{x}) \\
+\lambda_{2}\mathcal{L}_{CE}(\mathrm{y}, \mathrm{y}_{t}).
\label{generalinv_le_entropy}
\end{align}

\noindent
We used $\lambda_{1}$ = 0.02 and $\lambda_{2}$ = 0.1 in our experiments.

%% file: results.tex
\begin{figure*}
    \centering
    \includegraphics[width = 0.99\textwidth]{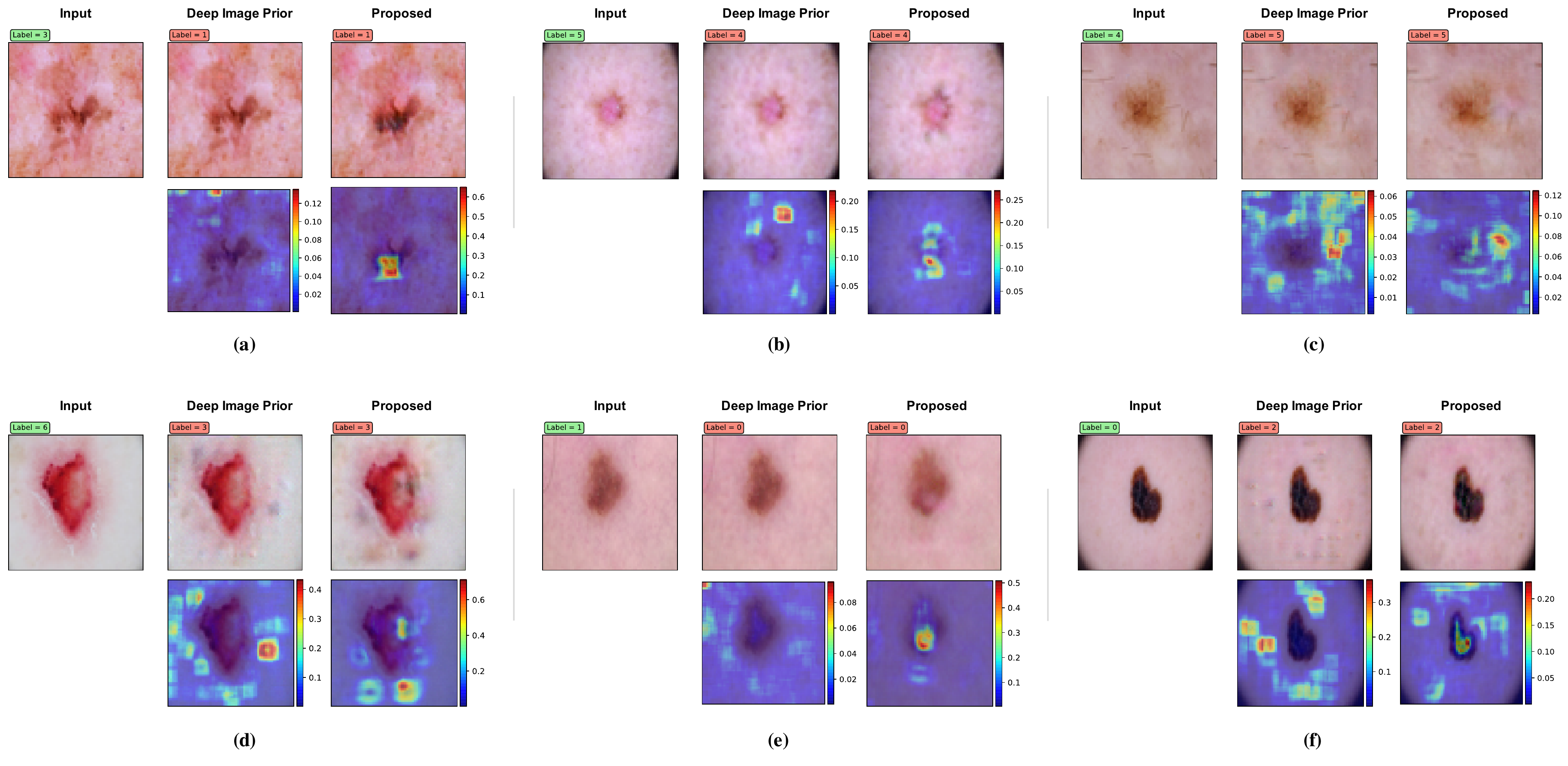}
    \caption{\textbf{Counterfactual Reasoning:} (a)-(f) Conventional DIP does not produce interpretable perturbations on the input and generates adversarial examples as counterfactuals (difference images show where the image has been perturbed). The proposed approach however introduces perturbations to the lesion specific regions, thus providing interpretable counterfactuals.}
    \label{fig:exp3}
\end{figure*}

In this section, we evaluate our proposed approach for pre-image recovery and counterfactual generation on the ISIC 2018 lesion diagnosis challenge dataset~\cite{codella2019skin, tschandl2018ham10000} containing a total of 10,015 dermoscopic lesion images, where each image is associated with one of the following 7 disease groups: Melanoma, Melanocytic nevus, Basal cell carcinoma, Actinic keratosis, Benign keratosis, Dermatofibroma and Vascular
lesion. We performed a stratified $90-10$ split on the dataset to train and evaluate the predictor and the loss estimator. In all experiments, we utilized a ResNet-18~\cite{he2016deep} model as the predictor while the loss estimator utilized the feature maps from the four residual blocks and subsequently transformed them using linear layers of $128$ hidden units. For DIP implementation, we employed a U-Net architecture with 4 downstream and upstream layers with skip connections and utilized the bilinear interpolation to perform upsampling in the upstream path. Following the standard DIP setup\footnote{https://github.com/DmitryUlyanov/deep-image-prior}, a random noise image is provided as input to the U-Net. For all pre-image and counterfactual generation experiments, we trained the U-Net based DIP model for 5000 iterations and considered $\Psi(.)$ as the first residual block of the predictor. 

\noindent
\textbf{Observations}: In order to provide a holistic reasoning to support our proposed approach, we performed two experiments to understand the performance of the DIP for pre-image recovery with and without domain shifts. In the first experiment, we performed pre-image recovery for image examples that lie in the manifold of the ISIC image data with and without the gradients from the loss estimator (we show the results only for the former in Figure \ref{fig:exp1}). While it is known that the standard DIP inversion already produces high-quality reconstructions, we find that including the loss estimator leads to marginal improvements ($\approx 0.5-1$dB). However, the benefits of the proposed approach are more apparent when we consider OOD samples. In this experiment, we carried out pre-image recovery for an out-of domain sample (random crop/blur on an image) with and without the gradients from the loss estimator. It can be observed from Figure \ref{fig:exp2} that DIP without gradients from the loss estimator overfits to the corrupted image while the proposed approach is able to effectively guide the optimization to recover a semantically meaningful image. 

This observation is particularly important when generating counterfactual explanations, wherein one needs to generate realizations from different class-conditioned distributions while incurring small changes to the input image. Not so surprisingly, as showed in Figure \ref{fig:exp3}, using the standard image prior leads to adversarial examples that propose imperceptible changes to irrelevant parts of the image (almost always to the non-lesion pixels), even though the predicted lesion type changes. In contrast, by enforcing the predictions to have a low loss value (i.e., increase the likelihood of being an in-distribution image), the proposed approach introduces systematic perturbation to the lesion pixels. Note that, these changes are highly consistent with the ABCD signatures (asymmetry, border, color, and diameter) adopted by clinicians for detecting lesion types in practice. While these counterfactuals can be utilized to study the behavior of the predictor model (e.g., decision boundaries, confidence regimes etc.), they can also be leveraged to forecast lesion severity and thereby enable practitioners to gain insights into the patterns that can potentially manifest for different lesions. 


%% file: conclusion.tex
In this paper, we proposed a novel regularization strategy for effectively guiding a Deep Image Prior to produce semantically meaningful pre-images, while inverting the internal representations of a black-box predictor. In particular, our approach is based on an auxiliary loss estimator, trained alongside with the predictive model, which captures the inherent model uncertainties, the semantics of the image manifold and can detect distribution shifts. Through extensive analysis, we showed that the gradients from the loss estimator helps in obtaining natural pre-images using DIP, even when the latent space encoding corresponded to an out-of-distribution example. We also showed that, such a regularizer is essential for producing realistic counterfactuals from the underlying data manifold for any user-specified hypothesis. In comparison with the conventional DIP, our proposed approach systematically introduces perturbations in the lesion specific regions, which corroborate strongly with the widely adopted signatures for lesion type detection.

%% file: main.bbl
\begin{thebibliography}{10}

\bibitem{ulyanov2018deep}
D.~Ulyanov, A.~Vedaldi, and V.~Lempitsky,
\newblock ``Deep image prior,''
\newblock in {\em Proceedings of the IEEE Conference on Computer Vision and
  Pattern Recognition}, 2018, pp. 9446--9454.

\bibitem{anirudh2020mimicgan}
R.~Anirudh, J.J Thiagarajan, B.~Kailkhura, and P.T. Bremer,
\newblock ``Mimicgan: Robust projection onto image manifolds with corruption
  mimicking,''
\newblock {\em International Journal of Computer Vision}, pp. 1--19, 2020.

\bibitem{shah2018solving}
V.~Shah and C.~Hegde,
\newblock ``Solving linear inverse problems using gan priors: An algorithm with
  provable guarantees,''
\newblock in {\em 2018 IEEE international conference on acoustics, speech and
  signal processing (ICASSP)}. IEEE, 2018, pp. 4609--4613.

\bibitem{yeh2017semantic}
R.A Yeh, C.~Chen, Teck Yian~L., Alexander~G. S., M.~Hasegawa-Johnson, and M.N
  Do,
\newblock ``Semantic image inpainting with deep generative models,''
\newblock in {\em Proceedings of the IEEE conference on computer vision and
  pattern recognition}, 2017, pp. 5485--5493.

\bibitem{asim2018blind}
M.~Asim, F.~Shamshad, and A.~Ahmed,
\newblock ``Blind image deconvolution using deep generative priors,''
\newblock {\em arXiv preprint arXiv:1802.04073}, 2018.

\bibitem{narayanaswamy2020unsupervised}
V.~Narayanaswamy, J.J Thiagarajan, R.~Anirudh, and A.~Spanias,
\newblock ``Unsupervised audio source separation using generative priors,''
\newblock {\em arXiv preprint arXiv:2005.13769}, 2020.

\bibitem{goodfellow2014generative}
I.~Goodfellow, J.~Pouget-Abadie, M.~Mirza, B.~Xu, D.~Warde-Farley, S.~Ozair,
  A.~Courville, and Y.~Bengio,
\newblock ``Generative adversarial nets,''
\newblock in {\em Advances in neural information processing systems}, 2014, pp.
  2672--2680.

\bibitem{mahendran2015understanding}
A.~Mahendran and A.~Vedaldi,
\newblock ``Understanding deep image representations by inverting them,''
\newblock in {\em Proceedings of the IEEE conference on computer vision and
  pattern recognition}, 2015, pp. 5188--5196.

\bibitem{dosovitskiy2016inverting}
A.~Dosovitskiy and T.~Brox,
\newblock ``Inverting visual representations with convolutional networks,''
\newblock in {\em Proceedings of the IEEE conference on computer vision and
  pattern recognition}, 2016, pp. 4829--4837.

\bibitem{yoo2019learning}
D.~Yoo and I.S Kweon,
\newblock ``Learning loss for active learning,''
\newblock in {\em Proceedings of the IEEE Conference on Computer Vision and
  Pattern Recognition}, 2019, pp. 93--102.

\bibitem{thiagarajan2020accurate}
J.J Thiagarajan, V.~Narayanaswamy, R.~Anirudh, P.T Bremer, and A.~Spanias,
\newblock ``Accurate and robust feature importance estimation under
  distribution shifts,''
\newblock {\em arXiv preprint arXiv:2009.14454}, 2020.

\bibitem{codella2019skin}
N.~Codella, V.~Rotemberg, P.~Tschandl, M.E. Celebi, S.~Dusza, D.~Gutman,
  B.~Helba, A.~Kalloo, K.~Liopyris, M.~Marchetti, et~al.,
\newblock ``Skin lesion analysis toward melanoma detection 2018: A challenge
  hosted by the international skin imaging collaboration (isic),''
\newblock {\em arXiv preprint arXiv:1902.03368}, 2019.

\bibitem{tschandl2018ham10000}
P.~Tschandl, C.~Rosendahl, and H.~Kittler,
\newblock ``The ham10000 dataset, a large collection of multi-source
  dermatoscopic images of common pigmented skin lesions,''
\newblock {\em Scientific data}, vol. 5, pp. 180161, 2018.

\bibitem{mahendran2016visualizing}
A.~Mahendran and A.~Vedaldi,
\newblock ``Visualizing deep convolutional neural networks using natural
  pre-images,''
\newblock {\em International Journal of Computer Vision}, vol. 120, no. 3, pp.
  233--255, 2016.

\bibitem{ronneberger2015u}
O.~Ronneberger, P.~Fischer, and T.~Brox,
\newblock ``U-net: Convolutional networks for biomedical image segmentation,''
\newblock in {\em International Conference on Medical image computing and
  computer-assisted intervention}. Springer, 2015, pp. 234--241.

\bibitem{he2016deep}
K.~He, X.~Zhang, S.~Ren, and J.~Sun,
\newblock ``Deep residual learning for image recognition,''
\newblock in {\em Proceedings of the IEEE conference on computer vision and
  pattern recognition}, 2016, pp. 770--778.

\bibitem{thiagarajan2020calibrating}
J.J Thiagarajan, P.~Sattigeri, D.~Rajan, and B.~Venkatesh,
\newblock ``Calibrating healthcare ai: Towards reliable and interpretable deep
  predictive models,''
\newblock {\em arXiv preprint arXiv:2004.14480}, 2020.

\end{thebibliography}
